# Towards Nation-wide Analytical Healthcare Infrastructures: A Privacy-Preserving Augmented Knee Rehabilitation Case Study

Boris Bačić[1,2,3][0000-0003-0305-4322], Claudiu Vasile[1][0009-0008-5837-3180], Chengwei Feng[1,2][0009-0002-4858-6226] and Marian G. Ciucă[4][0009-0006-7634-2594]

[1] AUT University, Auckland, New Zealand
[2] Institute of Biomedical Technologies (IBTec), Auckland, New Zealand
[3] Sports Performance Research Institute New Zealand (SPRINZ), Auckland, New Zealand
[4] Ovidius University of Constanta, Romania
`boris.bacic@aut.ac.nz`

**Abstract.** The purpose of this paper is to contribute towards the near-future privacy-preserving big data analytical healthcare platforms, capable of processing streamed or uploaded timeseries data or videos from patients. The experimental work includes a real-life knee rehabilitation video dataset capturing a set of exercises from simple and personalised to more general and challenging movements aimed for returning to sport. To convert video from mobile into privacy-preserving diagnostic timeseries data, we employed Google MediaPipe pose estimation. The developed proof-of-concept algorithms can augment knee exercise videos by overlaying the patient with stick figure elements while updating generated timeseries plot with knee angle estimation streamed as CSV file format. For patients and physiotherapists, video with side-to-side timeseries visually indicating potential issues such as excessive knee flexion or unstable knee movements or stick figure overlay errors is possible by setting a-priori knee-angle parameters. To address adherence to rehabilitation programme and quantify exercise sets and repetitions, our adaptive algorithm can correctly identify (91.67%–100%) of all exercises from side- and front-view videos. Transparent algorithm design for adaptive visual analysis of various knee exercise patterns contributes towards the interpretable AI and will inform near-future privacy-preserving, non-vendor locking, open-source developments for both end-user computing devices and as on-premises non-proprietary cloud platforms that can be deployed within the national healthcare system.

"A healthy person has a thousand wishes, a sick person just one." – the known proverb may come to mind when seeking rehabilitation after injury, stroke, recovering from surgery or managing conditions of repetitive Occupational Overuse

Syndrome (OOS), or other neurological conditions. Rehabilitation specialists typically diagnose and provide follow-up sessions with interventions, all intended to accelerate healing. As part of the intervention, there are mutual expectations of physiotherapist and patient, including commitment to a personalised home exercise programme, aimed at improving wellbeing, and shortening return-to-work or return-to-sport times.

## 1    Introduction

For a small country with just over 5 million people, New Zealand's "Physiotherapy services revenue is expected to rise at an annualised 2.7% over the five years through 2023-24, to total an estimated $519.2 million." [1]. As the steady increase in costs of medical and physiotherapy services is a global problem, the overarching question is: How can we use scientific and technological advancements to enhance rehabilitation and help to inform the design of next-generation nationwide analytical infrastructures that should not be relying on third-party offshore cloud services?

Ever since the concept of telerehabilitation in the 1990s [2] and Microsoft's Kinect exergames from 2010s, the ideas combining augmented video interaction to improve wellbeing through active movements, Internet videoconferencing and more recently popular exercise mirrors are not new. Hence, for this research the guiding question is: What related contexts, insights and artefacts from the presented case study on knee rehabilitation may contribute to the near-future technology advancements in adding value to rehabilitation processes and technology?

Considering the prospect of the next-generation big data analytical healthcare platforms able to capture and process video and data streams, the need for digital equality of end-users computing devices, the objectives of this research are aligned with the following research questions:

1. How can we use near-obsolete end-user computing devices to capture and optionally process rehabilitation data for promoting patients' compliance and adherence factors [3,2]?
2. To what extent Artificial Intelligence (AI) and Computer Vision (CV):
   - are needed for the reported knee rehabilitation case study?
   - could be applied to extended sets of knee exercises potentially transferrable to more general neuromotor rehabilitation for both end-user computing and healthcare platforms?

## 2    Methods and Materials

Combining sport science and AI methodologies, the three-phase multidisciplinary approach involves analysis and recommendations: (1) Considering the nature of injury, rehabilitation process, movement goals, and recommended physical rehabilitation exercises [4]; (2) Bridging biomechanics and AI by: identifying common properties of the movement, common errors [4], to be matched with features and output classes; (3) data processing aimed at analytical assessment and feedback for intended users.

### 2.1 Dataset Collection and Analysis

The collected dataset represents a partial video log of real-life right knee rehabilitation programme, capturing a range of exercises with occasional incorrect movements that are known as common errors (Fig. 1).
**Nature of injury**. The right knee was diagnosed with an injury resulting in inflammation and knee instability, without major damage requiring surgery.
**Common errors and properties of prescribed knee exercises.** The incorrect rehabilitating exercise movements are subject to qualitative assessment criteria of an expert and varying during the progression within the prescribed programme.

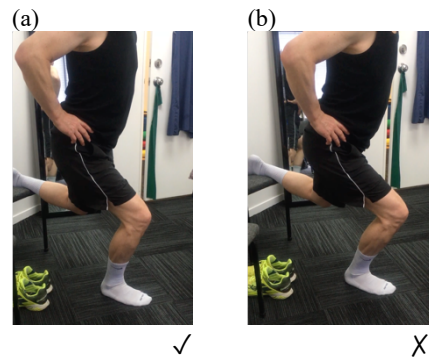

**Fig. 1.** Selected video frames showing correct posture and common errors from a physiotherapy session. Qualitative assessment criteria for (a) correct and (b) incorrect movements are based on evaluation parameters such as injured knee flexion (the knee is positioned in front of the foot index) and lateral instability i.e. tendency for the knee to collapse inwards.

**Data collection protocol.** All videos were recorded in full HD 1080p at 30 fps, using an iPhone (SE, iOS v.15.8.3). The camera was positioned approx. at waist level. The patient's right leg was either perpendicular to the camera (to estimate the knee angles in sagittal plane) or in frontal view to capture lateral knee instability. The collected video dataset (Table 1) contains different types of exercises (i.e. output classes or output labels) combining squat technique with both legs (S), single leg (SL), and Bulgarian split squat (BSS) as depicted in Fig. 1. To assist (A) balance and safety, some exercises were performed while holding a chair or a stick. To evaluate robustness of data processing and exercise counting functionality, most videos (Table 1) contain multiple exercises, including redundant elements such as waking into the scene, leg shaking off and knee lifting, which could be trimmed in future applications.

**Table 1.** Video dataset with summary of knee exercises.

| File name | Camera view | Exercises (output labels) | Total |
|---|---|---|---|
| IMG_1142.mp4 | Sagittal | 5 S | 5 |
| IMG_1143.mp4 | Sagittal | 1 S H, 6 SLA, 5 SL | 12 |
| IMG_1145.mp4 | Sagittal | 1 SA, 5 SLA, 1 S, 4 SL | 11 |
| IMG_1147.mp4 | Sagittal | 4 SA, 5 SLA, 5 SL | 14 |
| IMG_1150.mp4 | Sagittal | 5 S, 5 SLA, 5 SL, 1 SA, 7 SL, 8 S | 31 |
| IMG_2407.mp4 | Sagittal | 10 S, 2 S | 12 |
| IMG_2408.mp4 | Frontal | 7 S, 8 S, 8 S, 4 S | 27 |
| IMG_2409.mp4 | Sagittal | 5 SA, 1 BSSA, 9 BSSA, 2 SA, 3 SA, 8 BSS, 2 BSS, 5 BSS | 35 |
| IMG_2410_5.mp4 | Frontal | 10 S, 12 S, 10 S | 32 |
| | | Total exercises: | 179 |

**Rehabilitation Process and Movement Goals**. During the rehabilitation, the exercises evolved from simple and personalised to more general and challenging movements aimed to returning-to-sport demands.

From a sport science phasing analysis perspective [4], a common property of the movement for all prescribed right knee rehabilitation exercises is a spatiotemporal pattern starting with knee flexion and transitioning to knee extension.

From an AI perspective and preliminary scope of this work, it was decided that all exercise categories are to be merged into a single output class. Analytical assessment was provided as batch-processed feedback after exercising sessions. To promote adherence to home exercising programmes, the generated feedback should also include a quantitative summary of exercises performed. Considering potential barriers to digital inclusion (including access to the latest technology), video processing stages could be completed at the end-user side or performed on a near-future healthcare analytical platform. Options for privacy-preserving data exchange with near-future healthcare analytical platform are: MP4 video files, which could be filtered to preserve diagnostic information and/or CSV timeseries data.

### 2.2 Video Data Processing, Visualisation and Analysis

For prototyping purposes, there were three video data processing stages (Fig. 2):
1. **Video recording, post-data collection, transfer and conversion**. In view of broader end-user population, collected videos were transferred from iPhone SE (2016) into a MacBook Pro (2015) and PC computer (2012). All collected MOV video files were batch-converted into MP4 format using open-source ffmpeg command line utility (www.ffmpeg.org) as part of Linux bash shell script, achieving approx. compression ratio 1:5 or higher.

2. **Augmented video feedback, feature extraction and privacy-preserving numeric data generation**. For augmented replay, generated stick-figure elements of human pose estimation overlayed video with side-to-side live-plot.
3. **Providing exercise summary**. Generated CSV data table were pre-processed to filter rows containing NaN values. Knee angles were scaled and normalised for automated indexing, visualisation and exercise counting algorithm.

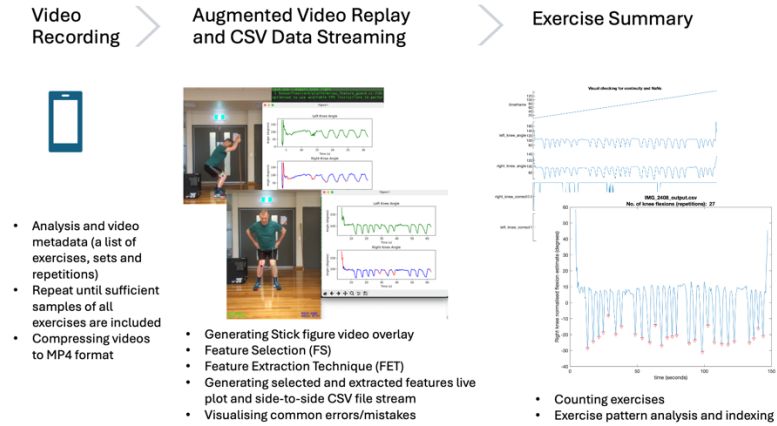

**Fig. 2.** Three-stage media processing: Video recording, augmented video replay, and exercise summary and indexing.

### 2.3 Algorithm Development and Visualisation

**Augmented Video Feedback and Feature Extraction**. Intended as a proof-of-concept and to enable further incremental development, we employed Google's MediaPipe for human pose estimation. From estimated 3D body landmarks, Feature Selection (FS) included hip, knee, ankle and foot index (tip of the shoe).

The key processing steps for augmented video replay include:
1. Frame-by-Frame Analysis: Each video frame is processed to detect body landmarks. If landmarks are detected, their coordinates are extracted.
2. Angle Calculation: The knee angles are computed using the coordinates of the hips, knees, and ankles. The script calculates the angle between the thigh and calf segments for both the left and right knees.
3. Safety Alerting Data Generation: In addition to generated timeseries data containing human pose estimation, and angle calculation, as part of Feature Extraction Technique (FET), we generated binary data indicating possible exercising errors.

**Front View and Side View Calculations**. The methodology presented distinguishes between frontal and side (sagittal) view calculations.

*Frontal view calculation insights*:
- For front view analysis (Fig. 3a), the algorithm incorporates the z-coordinates of the hips, knees, and ankles to calculate the angles in a 3D space. The hips (Fig. 3a), are denoted by A1 (left hip) and A2 (right hip), the knees by B1 (left) and B2 (right), and the ankles by C1 (left) and C2 (right). This approach ensures that the angles reflect the actual flexion and extension of the knees, even when the subject's legs move closer or further from the camera.

- The angle $\theta \equiv \angle ABC$ between the thigh and calf can be computed using the cosine rule (1) in a right-angled triangle:

$$\angle ABC = \cos^{-1}\left(\frac{BC^2 + AB^2 - AC^2}{2BC \cdot AB}\right) \quad (1)$$

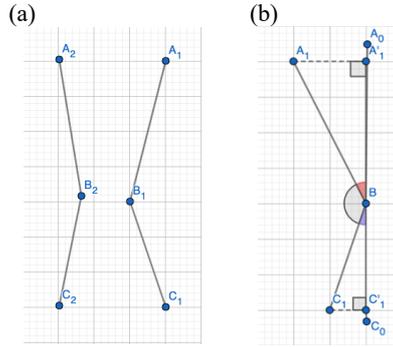

**Fig. 3.** Trigonometric conceptualisation of: (a) Frontal view of both knees and (b) Sagittal view of the right knee, where $A_0$ is the initial hip position, B the knee, and $C_0$ the ankle, while $A_1$ and $C_1$ are possible successive video frame estimations.

*Sagittal view calculation insights*:

- The algorithm calculates the angle between the thigh and calf segments using a 2D plane assumption (Fig. 3b). To compute the knee angles, the algorithm calculates the angle between the thigh and calf segments using the approximate body landmarks coordinates of the hips, knees, and ankles.
- The angles are adjusted based on the relative positions of the hips and ankles to account for any misalignment or oblique angles. The angle $\angle A_1BC_1$ between the thigh and calf (Fig. 3b) can be computed using the cosine rule (2) in a right-angled triangle or by applying the Law of Cosines (3):

$$\angle A_1BC_1 = \pi - \angle A_1BA_0 - \angle C_1BC_0 = \pi - \cos^{-1}\left(\frac{A'_1B}{A_0B}\right) - \cos^{-1}\left(\frac{C'_1B}{C_0B}\right) \quad (2)$$

$$\angle A_1BC_1 = \cos^{-1}\left(\frac{C_1B^2 + A_1B^2 - A_1C_1^2}{2C_1B \cdot A_1B}\right) \quad (3)$$

**Viewing Planes and Safety Alerting.** To highlight common errors, on video overlay and side-to-side timeseries plot (Fig. 2), we integrated customisable criteria for both front view and side view perspectives:

*Sagittal View Safety Alerting Rule Description.* The exercise is considered incorrect if the knee is positioned in front of the foot index.

*Frontal View Safety Alerting Rule Description.* The exercise is considered incorrect if the knee is positioned inwards relative to the foot, as determined using the collinearity condition ∆(*A*,*B*,*C*).

**Augmented video replay and CSV data stream generation.** The algorithm, depicted as a flowchart (Fig. 4) combines Computer Vision (CV) approaches such as Google MediaPipe for pose estimation and Feature Extraction Technique (FET) for augmented video replay and numeric timeseries data logging in CSV text files. Feature Extraction Technique (FET) included knee angle estimation and safety alerting binary data that may require expert's attention.

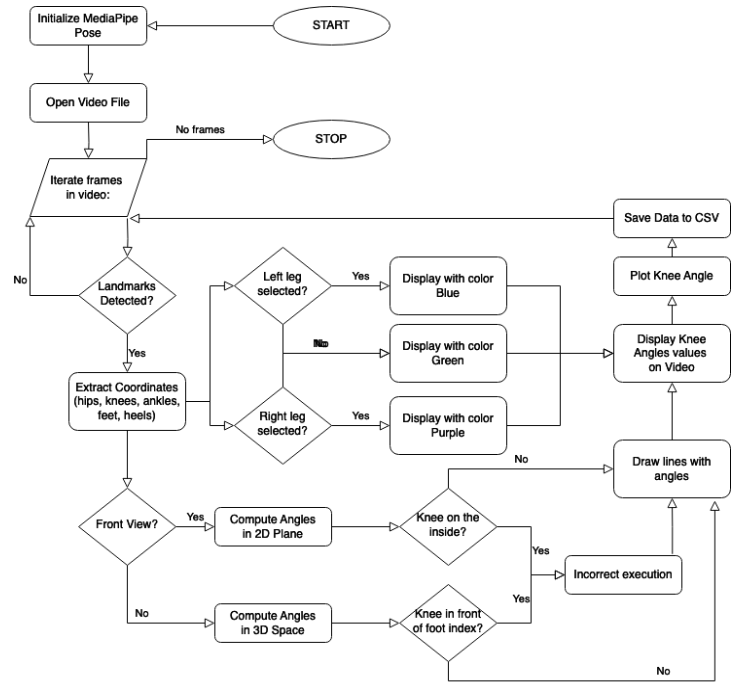

**Fig. 4.** Video processing flowchart of augmented video feedback.

**Exercise Summary: Counting and Indexing Visualisation Algorithm**. As the third stage of numeric data processing,
Algorithm **1** represents traditional computing approach, which the same as artificial neural networks (ANNs) would still require extracted features and initial knee exercise labelling information for its parameter tunning and validation. However, the algorithm did not require a larger labelled dataset for modelling numeric data timeseries.

**Algorithm 1**. Counting, indexing and visualising exercise variation patterns.

```
Input:  CSV_file, # Extracted pose estimation data and knee angles
        Sel_col   # List of selected columns, including right knee
Output: Knee_ex_cnt,                         # Exercise count
```

```
Preconditions: CSV_file, Sel_col           # Input data must exist
Default parameters (constants):
STD_TRESHOLD = 0.5                         # FP vs. FN ratio tunning
EXRC_FREQ    = 4  # Min. exercise time, filtering random movements
1:   T_knee_flex = read (CSV_file[Sel_col])
2:   T_knee_flex = preprocess(T_knee_flex)         # NaN removal
     ##  0-crossing and standard deviation timeseries analysis   ##
3:   Avg_knee_flex = mean(T_knee_flex[RightKneeAngle])
4:   for i = 1 to T_knee_flex[last] do
         T_knee_flex[i] = T_knee_flex [i] - Avg_knee_flex
     end do
5:   peakTreshold = standard_deviation(T_knee_flex) * STD_TRESHOLD
6:   [Knee_pks, Knee_locs] = findNegativePeaks(T_knee_flex, …
             MinPeakHeight=peakTreshold, MinPeakDistance=EXRC_FREQ)
7:   Knee_ex_cnt = length(Knee_pks)     # = size of Knee_pks array
8:   drawVisualisation(T_knee_flex, markers=[Knee_pks, Knee_locs])
9:   output (Knee_ex_cnt)
```

## 2.4 Results Analysis

Visual results (Fig. 5) achieved on rehabilitation case study dataset (Table 1), show color-coding of video overlays and side-to-side generated timeseries of knee angle stored as CSV numeric data.

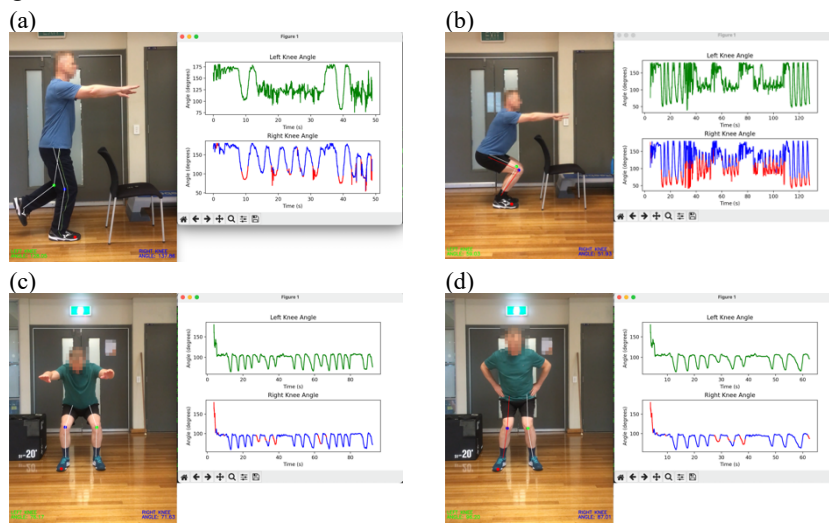

**Fig. 5.** Visual feedback results: (a) Correct posture (sagittal view), (b) potentially excessive knee angle where knee is positioned in front of the foot index, (c) correct posture (frontal view), and (d) incorrect posture, where the knees are collapsing inwards.

Safety alerting data (highlighted in red for the injured knee on Fig. 5c and Fig. 5d) is to draw expert or patient's attention to possible common errors. Exercise counting results achieved on rehabilitation case study dataset (Table 1), were tested on extracted CSV timeseries numeric data from individual video files.

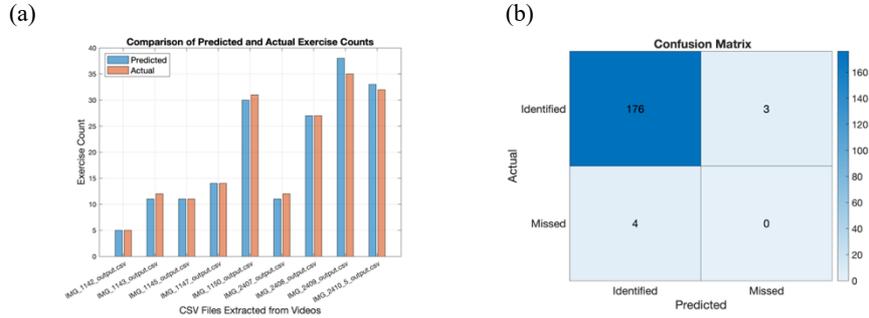

**Fig. 6.** Exercise counting and indexing results: (a) Per individual file and (b) overall.

Regarding robustness and adaptiveness, the default parameters remained the same for all processed video files (Fig. 6), including noisy data, random movements (e.g. shaking legs, lifting knees, and other loosening up movements) and mixed exercises repetitions.

## 3  Discussion

Generated post session safety-alerting data that may require the attention of a physiotherapy expert are recorded, however we did not yet implement real-time safety alerting functionality and mobile video streaming.

**The benefit of storing redundant data on 3D pose estimation with additional features**. CSV-formatted text files for post-exercise analysis, allow rapid prototyping and code base reusability and processing platform independence.

**Big-data Analytical Healthcare Platforms Considerations**. Aligned with the authors' views: (1) The experimental artefacts presented, transparent AI and algorithms are important for both end-users and healthcare cloud infrastructures; (2) national healthcare analytical platforms and infrastructure should operate within national systems and not rely on third-party cloud infrastructures based overseas, and; (3) the end-users should be in control of their privacy and choose what data they are willing to process and share.

## 4  Conclusion

In this research, we used real-life knee rehabilitation videos achieving a proof-of-concept of augmentation of video replay with extracted diagnostic information. The presented algorithms can be deployed on local near-obsolete computers or on near-future healthcare analytical platforms. Augmented video replay can show configurable overlay elements of stick figure, with generated side-to-side timeseries plot showing colour-coded safe and unsafe knee exercise positions. For local and cloud-based healthcare platform processing, privacy-preserving extracted numerical data (in CSV text format), contains human body pose estimation from video frames, and extracted exercise movement features to preserve diagnostic information. Adaptive exercise indexing algorithm can correctly identify (91.67%–100%) of all exercises from side- and front-view rehabilitation videos. For adherence to rehabilitation programme, counting and exercise indexing data for video log summary can also be used for additional diagnostic data expressed over time to inform decisions on personalised rehabilitation programme. Transparent algorithm design for adaptive visual analysis of various knee exercise patterns contributes towards interpretable AI objectives with logically acceptable and actionable feedback. For local/home-based and cloud/on-premises infrastructures, indexed data can: (1) Facilitate direct access to the movement or exercise of interest; (2) Be used to automatically trim or extract video sequence from the footage using open-source command-line utility such as ffmpeg.

Future work will be focused on: (1) Interpretable AI and near-future healthcare platforms based on horizontally scalable hardware and open-source software offering flexible privacy-preserving rehabilitation video and data logging; (2) augmented rehabilitation to be developed in real-time video streaming with safety alerting and enhanced video replays for local and cloud analytical platforms.

**Acknowledgement –** We wish to thank Auckland University of Technology (AUT) for providing access to the recording facilities and equipment. The first author wishes to thank Physiotherapy expert Suzanne McKenzie for her professional service, additional explanations, engaging practice and patience.

**Dataset Declaration –** The self-reported video dataset from the first author follows Auckland University of Technology Ethics Committee (AUTEC) guidelines (https://www.aut.ac.nz/research/researchethics/guidelines-and-procedures#6, accessed 21 November 2024) and it is aligned with exception section 6.4. "Research and teaching in which a single investigator is the subject of his/her own research and where no physically or psychologically hazardous procedure is involved."

## References

1. Gannon, D. (2023). Physiotherapy services in New Zealand - Market research report (2013-2028). https://www.ibisworld.com/nz/industry/physiotherapy-services/619/. Accessed 15 Nov. 2024.
2. Physiopedia (2020). Telerehabilitation and smartphone apps in physiotherapy. Accessed 20 Nov. 2024.
3. Physiopedia (2024). Adherence to home exercise programs. Accessed 20 Nov. 2024.
4. Knudson, D. V., & Morrison, C. S. (2002). Qualitative analysis of human movement (2nd ed.). Champaign, IL: Human Kinetics.